\documentclass[letterpaper, 10 pt, conference]{ieeeconf}  % Comment this line out
\IEEEoverridecommandlockouts                              % This command is only
\usepackage{booktabs}% For formal tables
\usepackage{todonotes}
\usepackage{subfigure}
\usepackage[noadjust]{cite}
\usepackage{algorithm}
\usepackage{amsmath}
\usepackage{algorithmic}
\usepackage{mcode}
\usepackage[hidelinks]{hyperref}
\usepackage{siunitx}
\usepackage{caption}
\usepackage{tabu}
\usepackage{multirow}
\usepackage{graphicx}
\title{\LARGE \bf Human Visual Attention Prediction Boosts Learning \&  Performance of Autonomous Driving Agents}

\author{Alexander Makrigiorgos, Ali Shafti, Alex Harston, Julien Gerard, and A. Aldo Faisal}

\begin{document}

\maketitle
\thispagestyle{empty}
\pagestyle{empty}

%%%%%%%%%%%%%%%%%%%%%%%%%%%%%%%%%%%%%%%%%%%%%%%%%%%%%%%%%%%%%%%%%%%%%%%%%%%%%%%%
\begin{abstract}
Autonomous driving is a multi-task problem requiring a deep understanding of the visual environment. End-to-end autonomous systems have attracted increasing interest as a method of learning to drive without exhaustively programming behaviours for different driving scenarios. When humans drive, they rely on a finely tuned sensory system which enables them to quickly acquire the information they need while filtering unnecessary details. This ability to identify task-specific high-interest regions within an image could be beneficial to autonomous driving agents and machine learning systems in general. To create a system capable of imitating human gaze patterns and visual attention, we collect eye movement data from human drivers in a virtual reality environment. We use this data to train deep neural networks predicting where humans are most likely to look when driving. We then use the outputs of this trained network to selectively mask driving images using a variety of masking techniques. Finally, autonomous driving agents are trained using these masked images as input. Upon comparison, we found that a dual-branch architecture which processes both raw and attention-masked images substantially outperforms all other models, reducing error in control signal predictions by 25.5\% compared to a standard end-to-end model trained only on raw images.
\end{abstract}
%%%%%%%%%%%%%%%%%%%%%%%%%%%%%%%%%%%%%%%%%%%%%%%%%%%%%%%%%%%%%%%%%%%%%%%%%%%%%%%%
\section{Introduction}
Autonomous driving is a rapidly expanding field of research with the potential to
provide enormous social, economic and environmental benefits to society. Current industry leaders in self-driving technology use a modular approach \cite{perceptionauv, ADdevelopment} which decomposes the driving task into multiple sub-modules such as perception, path planning and control. One alternative to this method which has seen increased interest in recent years due to the widespread popularity and effectiveness of deep neural networks is end-to-end imitation learning \cite{Pomerleau, Bojarski, Codevilla, Fusion, deepdriving, DAVE}, which attempts to directly learn to produce control signals from sensory input such as RGB camera images. In theory, these networks can continuously optimize their performance by training on unlabeled data collected from human-operated vehicles, without needing to explicitly define any road rules or hard-code specific driving behaviours. In practice, however, these networks can be unpredictable and have difficulty reacting to the presence of multiple dynamic objects \cite{limitations}. In this work, we aim to improve these driving agents by taking inspiration from the way in which humans approach the driving task.

Humans have a finely tuned sensory system which allows them to quickly locate objects of visual interest within a scene, ignoring irrelevant details and only processing what is important \cite{25yrs, Henderson2018-zg}. Eye movements have been shown to encode information about humans' intentions, understanding of their environment and decision-making processes \cite{Yarbus1967, kowler1990eye, 25yrs, perceptual_strats, tea, steer, Pekkanen2018-ez}, and have thus been used successfully as intuitive human-robot interfaces \cite{tostado20163d,maimon2017towards,shafti2019gaze, Lethaus2013-pi}. The ability to understand and recreate this attention-focusing mechanism has the potential to provide similar benefits to machine learning systems which have to parse complex visual scenes in order to complete their tasks, reducing computational burden and improving performance by quickly identifying key features in images. We attempt to demonstrate this benefit by incorporating an attention model which has learned to predict human gaze patterns while driving into the learning process of an end-to-end autonomous driving agent. By directing the focus of the driving network to specific image locations which contain important information for driving, we look to improve both the training speed and overall performance of these agents.

\begin{figure}[tp]
\includegraphics[width=\columnwidth]{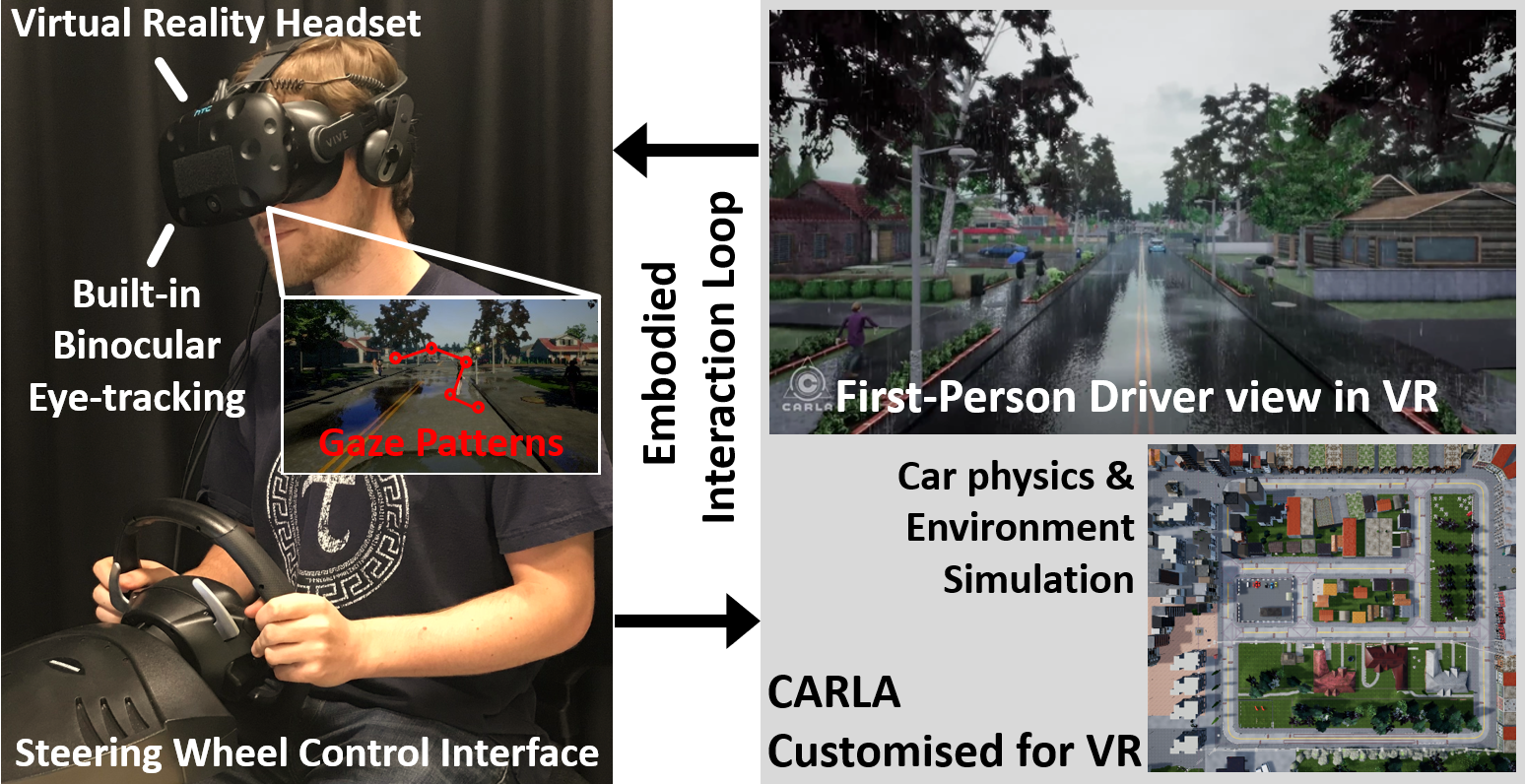}
\caption{A subject driving in the VR simulation.}
\label{system}
\end{figure}

Few attempts have been made thus far to incorporate attentional mechanisms into the training of autonomous vehicles, despite the success of these mechanisms for other applications \cite{ShowAttendTell, Attention, captioning, AGIL,planningautonomousvehic}. In \cite{Auxiliary}, a set of auxiliary predictions related to the driving task are defined and incorporated into the training process of an end-to-end network with the intention of guiding the agent towards useful driving features. Our method aims to inject human knowledge of the driving task into these agents in a more natural and data-driven way, by training a visual attention model directly on human eye movement data and using the outputs of this network to augment the subsequent end-to-end training of an autonomous driving agent.

In Section II, we describe our gaze data collection experiments and the details of the deep neural network which we use to predict saliency for this task. Section III presents our methods for incorporating the saliency maps produced by our gaze network into the training of self-driving vehicles. Our quantitative evaluations of both systems are reported and discussed in Section IV. Section V concludes the paper and discusses directions for future work.
%%%%%%%%%%%%%%%%%%%%%%%%%%%%%%%%%%%%%%%%%%%%%%%%%%%%%%%%%%%%%%%%%%%%%%%%%%%%%%%%
\section{Gaze Prediction}
Our goal is to create a system which can imitate human visual attention dynamics in order to improve the performance of an autonomous driving system. To accomplish this, we collect gaze data from human drivers in the CARLA driving simulator \cite{CARLA} and train a deep neural network to predict gaze attention on unseen images. In this section, we describe our data collection procedure and the details of the gaze prediction model.
%\subsection{System overview and architecture}
%Our goal is to create a system which can imitate human attention dynamics in order to improve the %performance of an autonomous driving system. The system trains in two main stages. First, a gaze %network is implemented which predicts where humans will look in a driving scene. To collect %high-quality human gaze data in a driving environment, we use a version of the CARLA %simulator\cite{CARLA} which has been modified to run in a virtual reality environment and track the %location of the user's gaze in terms of 3D CARLA coordinates. We use forward projection to convert the %3D gaze coordinates to 2D pixel coordinates corresponding to the proper location in images recorded by %a vehicle-mounted camera, and train our gaze network on these images.

%After fully training the gaze network, we use it to pre-compute visual attention maps for a large %dataset of driving images produced by an expert autopilot agent provided by the CARLA developers. We %then train an autonomous driving agent using a dual-branch architecture which learns from both raw and %attention-masked images. Figure \ref{architecture} shows the full architecture of our system; in the %following subsections, we describe each component in detail.

 \subsection{Gaze Data Collection}
 Our experiments are conducted using the CARLA driving simulator \cite{CARLA}, an increasingly popular open-source environment for training and evaluating autonomous driving systems. To create as realistic of a driving environment as possible, we modified CARLA to run in virtual reality using SteamVR \cite{CARLAVR}. Subjects control the vehicle using a Ferrari 458 Italia USB steering wheel controller which communicates with the CARLA server via a Python client -- a picture of our experimental setup can be viewed in Figure \ref{system}. For each subject, we conduct six consecutive three-minute driving episodes in CARLA's `Town01' environment, varying weather settings between each episode and using a fixed framerate of 25 fps. Subjects are instructed to drive safely and respect all road rules, but are otherwise given no directions or target destinations. A total of 243,000 frames, corresponding to 2.7 hours of driving data, was collected.
 
 To track the subject's eye movements, we use an HTC Vive headset with built-in SMI Gaze Tracking sensors, recording the absolute location of the user's gaze in (X,Y,Z) CARLA coordinates at each frame. The subject may freely view their environment by moving their head, but for the purposes of our gaze prediction and autonomous driving models, we require a fixed point of view. We therefore use a forward-facing camera mounted on the roof of the vehicle to record RGB images at each frame. Finally, we record the camera-to-car and car-to-world transformation matrices provided by the CARLA API in order to map between World and Pixel coordinates, as described below.
 
 %We perform forward projection to retrieve the 2D pixel location of the user's gaze in the recorded %RGB images from the absolute 3D CARLA coordinates provided by the gaze tracking system. First, we %transform the pixels from the World coordinate system to the Camera coordinate system, in which %the camera is always located at the origin, with the lens pointed down the Z axis, and the X and Y %axes defining the image plane. This is a rigid 3D transformation which requires knowledge of the %rotation and translation of the camera relative to the world, referred to as a camera's %\textit{extrinsic} matrix $M_{ext}$. CARLA directly provides the transformation between any %cameras and the vehicle it is attached to, as well as the transformation between the player's %vehicle and the world. The extrinsic matrix is thus easily obtained by multiplying these two %matrices and inverting the resulting camera-to-world transform. 
 
We perform forward projection to retrieve the 2D pixel location of the user's gaze in the recorded RGB images from the absolute 3D CARLA coordinates provided by the gaze tracking system. At each time step, CARLA directly provides a matrix which describe the 3D transformation between the Camera and World coordinate systems, referred to as the camera's \textit{extrinsic} matrix $M_{ext}$. We then use perspective projection to convert from 3D Camera coordinates to 2D Image coordinates, projecting our 3D gaze locations onto an X-Y plane located $f$ pixels away from the camera's origin on the Z axis and scaling them to the resolution of our images $(W,H)$. The projection and scaling operations can be combined into a single matrix, known as the camera's \textit{intrinsic} matrix $M_{int}$, and multiplied by the 3D camera coordinates of the user's gaze to retrieve the final pixel coordinates. In cases where the user looked outside the field of view of the camera, the returned pixels are not within the image boundaries and the frame is marked as empty of fixations (1.36\% of frames). The forward projection process is summarized in Equation \ref{forwardprojection}. To verify the correctness of these calculations, we performed a test in which the SMIGazeCursor object used to report gaze locations in the Unreal Engine was made visible as a red sphere in the CARLA environment, thus appearing in the recorded camera images, and visually confirmed that the object was appearing in the same (x,y) locations produced by our calculations.

 \begin{equation}
     \begin{bmatrix}
     x \\
     y \\
     1 \\
     \end{bmatrix}
     = 
     \underbrace{
     \begin{bmatrix}
     f & 0 & W/2 & 0\\ 
     0 & f & H/2 & 0\\
     0 & 0 & 1 & 0\\
     \end{bmatrix}
     }_\text{$M_{int}$}
     \underbrace{
     \begin{bmatrix}
     r_{1,1} & r_{1,2} & r_{1,3} & t_1 \\
     r_{2,1} & r_{2,2} & r_{2,3} & t_2 \\
     r_{3,1} & r_{3,2} & r_{3,3} & t_3 \\
     0 & 0 & 0 & 1 \\
     \end{bmatrix}
     }_\text{$M_{ext}$}
     \begin{bmatrix}
     X\\
     Y\\
     Z\\
     1\\
     \end{bmatrix}
 \label{forwardprojection}
 \end{equation}
 
\begin{figure}[tp]
\includegraphics[width=\columnwidth]{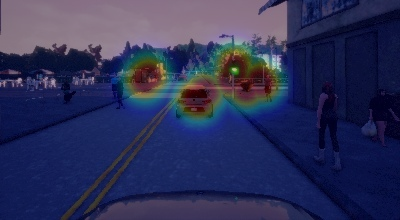}
\caption{Example fixation map generated from ground truth gaze data on a driving image from our dataset, containing all fixations within 1 second of the current frame.}
\label{attentionmap}
\end{figure}

 We compute continuous attention maps from gaze fixations using the following method suggested by \cite{Dreyeve}. To construct a fixation map $F_t$ for a frame at time $t$, we first retrieve the 3D gaze locations from time steps $t-12$ to $t+12$, for a total of 25 fixations (including the current frame), representing a 1-second window of time centered on $t$. These absolute locations are then converted to x,y coordinates within the current frame using the described forward projection process, with all (X,Y,Z) gaze locations transformed using the extrinsic matrix for the current frame. A multivariate Gaussian is then centered on each pair of pixel coordinates, and the 25 Gaussians are combined into a single attention map via a $max$ operation, which is finally normalized, resulting in a continuous probability distribution modeling the subject's visual attention for the current frame. An example of a driving image blended with its generated attention map is shown in Figure \ref{attentionmap}.

A sizeable portion of the collected data consists simply of periods where the driver is waiting at a traffic light, as the CARLA environment simulates a small urban town. To distinguish between active driving sequences, where the driver must pay constant attention to avoid collisions or drifting from their lane, and traffic lights where minimal attention must be paid (and thus eye movements may be less predictable), a manual annotation of the dataset was performed to label all frames as either 'driving' or 'traffic'. Of the 243,000 recorded frames, 79,437 (32.7\%) are labeled as 'traffic', with the remaining 163,563 (67.3\%) labeled as 'driving'. In the Experimental Evaluation section, we see that our gaze network performs significantly better on driving sequences.

\subsection{Network Architecture}
Our gaze prediction model is based on the DR(eye)VE model of \cite{Dreyeve}, combining predictions from separate branches trained on different information domains to output full-resolution attention maps. The overall model is composed of two identical saliency branches, the first of which processes sequences of raw RGB images, and the second of which processes sequences of dense optical flow maps which have been precomputed for our entire dataset using the Farneback optical flow algorithm \cite{Farneback}. The two branches are first trained independently, and subsequently fine-tuned by training both branches simultaneously, such that the network learns to weight the contributions of each branch and optimize its performance. 

The individual architecture of each saliency branch consists of two training streams, each taking a different input and producing their own outputs. Given an input tensor with N frames of size $X_{orig}$, $Y_{orig}$ , the first stream simply takes this tensor, resizes the images to $X_{res}, Y_{res}$, and processes it with a 3D convolutional network which produces a fixation map of size $X_{res}, Y_{res}$. This network is referred to as the COARSE module, and is strongly based on the C3D architecture of Tran et al. \cite{C3D}. The resulting fixation map is then upsampled to the original size $X_{orig}, Y_{orig}$, concatenated with the final frame of the input sequence, and further refined by a series of 2D convolutional layers (the REFINE module), outputting a full-sized saliency map. The second stream, rather than resizing the input images, takes as input a randomly cropped section of the images, also of size $X_{res}, Y_{res}$. It then processes this random crop using the same COARSE network as the first stream, with shared weights, and directly outputs a saliency map prediction for the area which was chosen by the random crop. The full-frame and cropped losses are computed separately for the two streams, and summed to calculate the final loss for the branch.  This random cropping strategy can be thought of as a form of data augmentation which forces the network to pay attention to features within the image rather than simply the spatial location of the labels; the majority of the cropped areas are outside of the center of the image, and by including the loss from these predictions in the branch's overall loss, the network is penalized if it begins to overfit to the mean of the dataset and predict only in central regions of the input images. The loss for each attention map is calculated as the Kullback-Liebler Divergence between the ground truth ($Y$) and predicted ($\hat{Y}$) maps for all pixels $i$: 

\begin{equation}
D_{KL}(Y\lvert\rvert\hat{Y}) = \sum_{i}Y(i) log \left( \epsilon + \frac{Y(i)}{\epsilon + \hat{Y}(i)}\right)
\end{equation}

where $\epsilon$ is a small regularization constant.

We make one notable addition to the model of \cite{Dreyeve}. Drawing inspiration from the branched architecture of the Conditional Imitation Learning (CIL) networks which we use to train our autonomous driving agents \cite{Codevilla}, we introduce the notion of a high-level command into the gaze network which represents the current intention of the driver at each frame. Our commands correspond exactly to the four commands used to control autonomous driving agents' behaviours in the CIL framework: ``Follow Lane", ``Turn Left", ``Turn Right", and ``Go Straight". We hypothesized that giving the Gaze network access to this signal that represents the driver's intentions for a given input clip will lead to more accurate predictions about where the driver will look. This is a logical assumption given the top-down nature of the task; a driver intending to make a left turn at the upcoming intersection, for example, will likely spend more time inspecting the left lane for oncoming traffic or pedestrians. As this intention signal is already being provided for the autonomous driving agent which the gaze network is intended to augment, it can be easily provided to the gaze network as well. We manually labeled our dataset with these high-level commands and trained a separate saliency prediction branch for each command. When precomputing attention maps for our driving dataset frames, our intention-branched model switches which branch is used to make predictions based on the high-level command recorded by the expert demonstrator for those frames. We refer to this model as the ``Intention-Branched DR(eye)VE" model.

\begin{figure*}[!t]
\includegraphics[width=2\columnwidth]{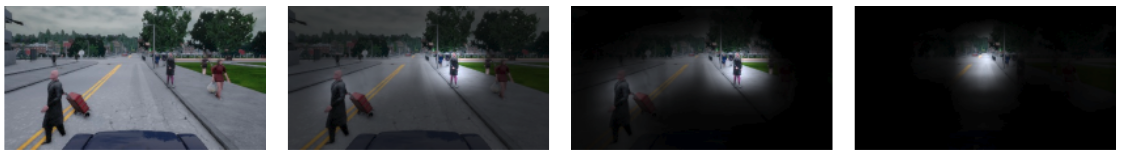}
\caption{Example driving image with different types of attention masking applied. From left to right: Raw image, Soft masking, Hard masking, Baseline masking. Hard and soft masks use the attention map generated by our gaze prediction network, while the baseline mask uses the mean fixation map from our gaze network's training dataset.}
\vspace{-15pt}
\label{maskingexample}
\end{figure*}

\section{Autonomous Driving Agents}

\subsection{Data collection}
We use an automated navigation expert provided by the CARLA developers\footnote{\url{https://github.com/carla-simulator/data-collector}} to collect data for the training of our autonomous driving agents. This agent has access to privileged information about the state of the server at all times, such as a map of its environment and the exact positions of all other agents in the simulation. Using this information along with hard-coded rules about how to react in traffic situations (e.g. stop at red traffic lights or when pedestrians are ahead), the vehicle navigates from an initial random position to a randomly chosen destination, recording images from a camera mounted in the same position as in our gaze collection experiments along with the driving signals (throttle, brake, steering angle, current speed) used to train the networks and the high-level commands associated with each image. Upon reaching the destination, the episode is terminated and a new episode begins, randomizing the weather settings and the number and location of pedestrians and other vehicles. Noise is occasionally added to the agent's actions to increase the diversity of the training set, but this addition is not reflected in the recorded commands to ensure that trained agents do not learn noisy behaviour. In the rare case that the noise results in a collision, the episode is terminated and is not used for training purposes. 

Data is collected from both towns existing in the stable version of CARLA -- `Town01' and `Town02', with the training set composed only of `Town01' images and the test set containing images from both towns. Weather presets were randomly chosen from the `Clear Noon', `Cloudy Noon', `Mid Rainy Sunset', `Hard Rainy Noon', `Wet Noon', and `Clear Sunset' presets. The number of pedestrians spawned ranged from 120 to 200, and the number of vehicles from 20 to 40. Episodes typically ranged between 1000 and 8000 frames, with 229 total episodes resulting in 657,456 frames, equivalent to 7.3 hours of driving at 25 frames per second.

\subsection{Attention masking}
Rather than compute saliency maps on the fly while training an imitation learning network, which would drastically slow down the training process and require computing the same attention maps many times, we pre-compute the saliency maps for all frames in the training and test sequences using our fully trained gaze prediction network. As mentioned in the Gaze Prediction section, the branch of our gaze network used to predict the saliency map for each frame was switched based on the high-level command recorded for that frame by the navigation expert. The computed maps are then used to mask the images from which they were sampled. Three different types of masks are generated, which we refer to as hard attention, soft attention, and baseline masks. Hard attention masks are simply computed via element-wise multiplication (denoted by $\odot$) of the attention map $F_{t}$ and the input camera image $I_t$, blacking out any regions which were not predicted likely to contain fixations by the gaze network:

\begin{equation}
M_{hard} = I_t \odot F_t
\end{equation}

Soft attention masks are computed by reducing the brightness of the images by a factor of $\lambda$ in areas where the attention map does not overlap with the image:

\begin{equation}
M_{soft} = \lambda*I_t + (1-\lambda)*I_t \odot F_t
\end{equation}

Finally, baseline masks are computed via element-wise multiplication of the camera images with the mean fixation map of our gaze dataset $G$:

\begin{equation}
M_{base} = I_t \odot G
\end{equation}

Figure \ref{maskingexample} shows an example of each type of masking applied to the same driving image.

\subsection{Driving agents}
\vspace{-25pt}
To train our imitation learning agents for autonomous driving, we use the Conditional Imitation Learning (CIL) framework of \cite{Codevilla}. In this framework, the driving agent receives two inputs: an RGB camera image and a float indicating the current speed of the vehicle. The RGB image is processed by a Convolutional Neural Network, while the speed input is processed by two fully-connected layers. These outputs are then concatenated, processed by two more fully connected layers, and fed into one of five branch heads, selected based on the high-level command associated with the input frame. Each branch head is associated with one of four commands: Follow, Left, Right, and Straight, with the fifth branch used in the rare case that no high-level command is provided (this sometimes occurs for a small number of frames when the expert demonstrator has reached its target destination but the episode has not yet terminated). All branch heads are formed of two fully-connected layers, and output four floats representing the actions for the Steer, Throttle and Brake commands and the predicted current speed of the vehicle. The speed input and output signals are necessary to produce sensible acceleration or braking commands in the presence of dynamic objects. 

\begin{figure*}[!t]
\includegraphics[width=1.8\columnwidth]{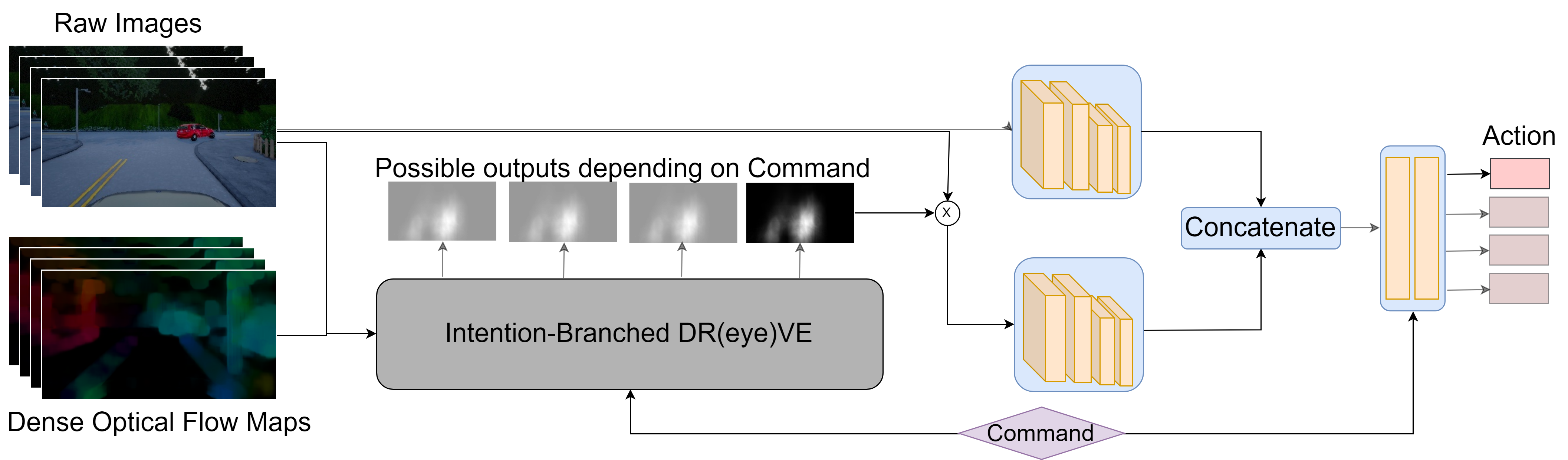}
\caption{Architecture for the dual-branch driving agent which is trained on both raw and hard-masked mages.}
\label{dualbranch}
\vspace{-15pt}
\end{figure*}
\begin{figure}[!bp]
\includegraphics[width=\columnwidth]{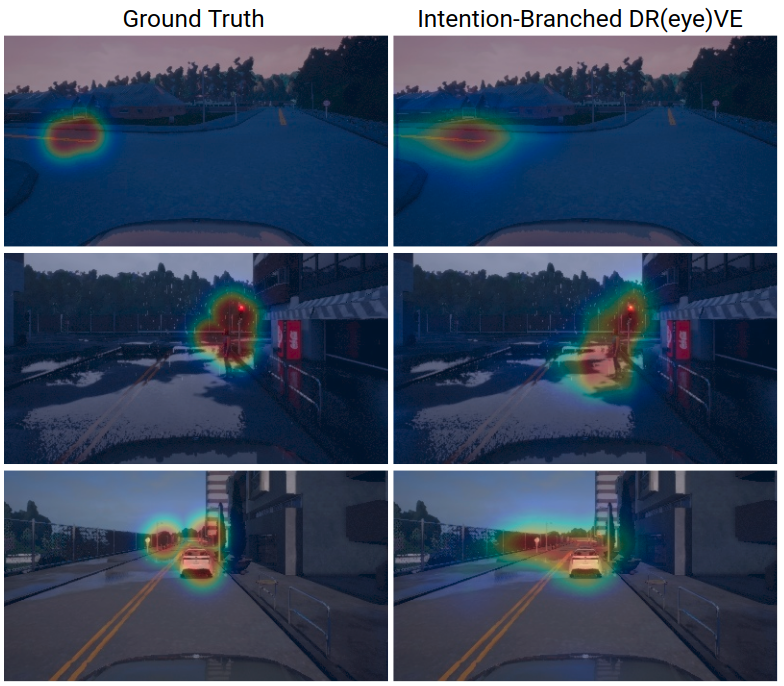}
\caption{Three sample outputs from our Intention-Branched DR(eye)VE model compared to the ground truth fixations of the human driver.}
\label{ibdpreds}
\end{figure}

We train and evaluate a total of five different driving agents on the images from our driving dataset. Four of these agents use the exact CIL architecture described above; of these four, one is trained only on raw images, one on hard-masked images, one on soft-masked images, and one on baseline-masked images. We refer to these agents henceforth as the raw, hard, soft, and baseline agents respectively. The fifth agent uses a dual-branch architecture inspired by \cite{AGIL}, with the top branch taking raw images as input and the bottom branch taking hard-masked images. This agent is referred to as the dual-branch agent, and is depicted in Figure \ref{dualbranch}. Each branch uses the same convolutional architecture as the CIL models, and the output features are concatenated before being processed by two fully connected layers and branching on the current high-level command, just as in the CIL architecture. 

To ensure a fair comparison between models, all agents are trained with the same parameters, for an equal number of steps and using the same loss function. We use the loss function defined in \cite{Codevilla}, which weights and sums the mean squared error between each predicted control value (throttle, brake, steering angle, speed) and the values used by the navigation expert. 

\section{Results and Discussion}
\subsection{Gaze network results}
 Based on saliency metric suggestions in \cite{saliencymetrics}, we evaluate the performance of our gaze network in terms of the Kullback-Liebler Divergence ($D_{KL}$) and Correlation Coefficient ($CC$) between ground truth and predicted saliency maps. Our model's scores are compared to those of three bottom-up saliency models which have achieved good performance on saliency benchmarks such as MIT300 and SALICON: DeepGaze II \cite{DG2}, MLNet \cite{MLNet}, and RMDN \cite{RMDN}, as well as the DR(eye)VE \cite{Dreyeve} model which our model attempts to improve on. The MLNet, RMDN and DR(eye)VE models are retrained from scratch on our dataset, while for DeepGaze II we use a pretrained model\footnote{acquired from \url{https://deepgaze.bethgelab.org}}. All models are evaluated on six driving episodes (27,000 frames) which were not included in the training data, with the results for both the full test episodes and the driving frames reported in Table \ref{tab:gaze}.

 Our model outperforms all other saliency models in both metrics, showing a major improvement over the bottom-up models as expected and slightly improving upon the original DR(eye)VE network, validating our intention-branched design. We also note a substantial improvement in prediction accuracy between the full test episodes and the driving frames, confirming our assumption that eye movements are more difficult to predict when the driver is sitting in traffic and does not necessarily need to pay attention to the road. Indeed, it seems to be the case that in traffic frames, bottom-up features play a larger role in attracting gaze, as indicated by the improved performance of the pre-trained bottom-up DeepGaze II network on the full episodes which include traffic sequences. This is a positive finding, as we are not overly concerned with the model's predictions while sitting still in traffic since no complex driving actions are executed in these scenarios.
 
 Overall, we find that our model appears to have learned a solid representation of human attention dynamics for the driving task, consistently paying attention to important objects such as other vehicles, lane markings, traffic lights and nearby pedestrians while ignoring irrelevant objects which distracted the bottom-up networks such as Coca-Cola machines, trees or mailboxes. Figure \ref{ibdpreds} compares some of our model's predictions for frames in the test set against the ground truth fixations of the human driver. A longer driving sequence recorded by the expert navigation agent and blended with our model's saliency predictions can be viewed in our video: \url{https://youtu.be/X3Xz7GMQX3s}.
 
\begin{table}[bp]
\captionsetup{justification=centering}
\caption{Results for five gaze prediction models evaluated on our test episodes. We compare performance on both the full test set and the subset of frames which were labeled as driving frames.}
\label{tab:gaze}
\resizebox{\columnwidth}{!}{%
\begin{tabular}{|c|c|c|c|c|}
\hline
& \multicolumn{2}{c|}{Driving + Traffic Frames} & \multicolumn{2}{c|}{Driving Frames} \\
\hline
\textbf{Network} & $\mathbf{D_{KL}} (\downarrow)$ & $\mathbf{CC} (\uparrow)$ & $\mathbf{D_{KL}} (\downarrow)$ & $\mathbf{CC} (\uparrow)$ \\ \hline
\textbf{DeepGaze II} \cite{DG2} & 1.287 & 0.511 & 1.340 & 0.503 \\
\hline
\textbf{MLNet} \cite{MLNet} & 1.222 & 0.553 & 1.183 & 0.570 \\
\hline
\textbf{RMDN} \cite{RMDN} & 1.029 & 0.600 & 1.073 & 0.592 \\
\hline
\textbf{DR(eye)VE} \cite{Dreyeve} & 0.654 & 0.761 & 0.600 & 0.787 \\
\hline
\textbf{\begin{tabular}[c]{@{}c@{}}Intention-Branched \\ DR(eye)VE (ours)\end{tabular}} & \textbf{0.642} & \textbf{0.763} & \textbf{0.589} & \textbf{0.790} \\
\hline

\end{tabular}%
}
\end{table}

\subsection{Driving agent results}
We perform offline evaluation of our driving agents on a test set consisting of 150,000 previously unseen frames and control signals provided by the expert navigator. In order to track both learning speed and overall performance, we evaluate each agent's performance on the full test set every ten thousand training steps. We report our final results in terms of both the task-weighted Mean Squared Error (MSE), which is used as a loss function to train the agents, and the task-weighted Mean Absolute Error (MAE) for all four tasks, which has been shown to have a stronger correlation with real-world driving performance \cite{offlinemetrics}. 

\begin{figure}[tp]
\includegraphics[width=\columnwidth]{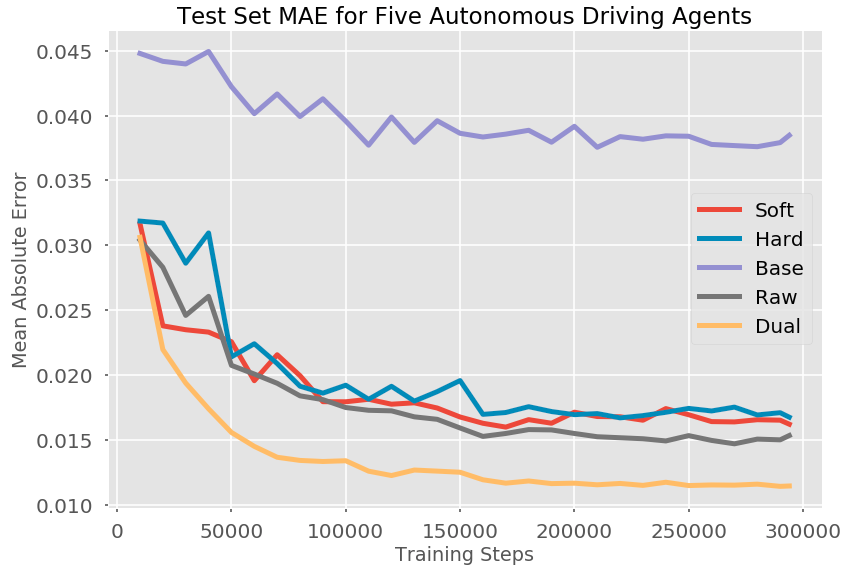}
\caption{Test set MAE for all driving agents, evaluated throughout the training process. The dual-branch agent shows a pronounced improvement over all other agents for this metric, which has been shown to correlate more strongly with real-world driving performance than MSE \cite{offlinemetrics}.}
\label{drivemae}
\end{figure}

The test set scores in these two metrics for our five fully trained driving agents, averaged across three training runs for each agent, are reported in Table \ref{tab:drive}. The dual-branch model achieves the minimum prediction error for both metrics by a considerable margin, with an 12.6\% decrease in MSE and a 25.5\% decrease in MAE compared to the predictions of the raw agent. In Figure \ref{drivemae}, we show the test set MAE of all agents evaluated throughout the training process. In addition to achieving better overall performance, the dual-branch agent learns much more rapidly than all other agents, attaining a lower MAE in sixty thousand training steps than any other model achieves even when fully trained. 

There is very little difference in prediction error between the raw, soft and hard agents. Our expectation was that the soft agent would outperform the raw agent, as it contains the attention information produced by our Intention-Branched DR(eye)VE model while also retaining all of the original image data. However, the agent does not appear to take advantage of the attention information when it is presented in this way, and is in fact marginally outperformed by the raw agent. The difference between the soft agent and the dual-branch agent, both of which theoretically have access to the same information, highlights the importance of model architectures when incorporating attention to machine learning systems. Equally unexpected was the discovery that the hard agent performed approximately the same as the soft and raw agents, given the extent of the information which is blacked out during the image masking process. The baseline and hard agents had similar amounts of information removed by the hard masking process, but while the baseline agent was severely affected by this removal, the hard agent continued to perform well due to the selective nature of the masking. This is a compelling affirmation of our Intention-Branched DR(eye)VE model's ability to correctly identify the regions of an image which are important to the driving task.

\begin{table}[tp]
\captionsetup{justification=centering}
\caption{Test set prediction errors for five autonomous driving agents trained using the attention maps produced by our gaze network.}
\label{tab:drive}
\resizebox{\columnwidth}{!}{%
\begin{tabular}{|c|c|c|}
\hline
\textbf{Agent} & \textbf{Multi-Task MSE} $(\downarrow)$ & \textbf{Multi-Task MAE} $(\downarrow)$ \\
\hline
\textbf{Raw} \cite{Codevilla} & 0.0143 & 0.0153 \\
\hline
\textbf{Baseline} (ours) & 0.0381 & 0.0385 \\
\hline
\textbf{Hard} (ours) & 0.0146 & 0.0168 \\
\hline
\textbf{Soft} (ours) & 0.0142 & 0.0162 \\
\hline
\textbf{Dual-branch} (ours) & \textbf{0.0125} & \textbf{0.0114} \\
\hline

\end{tabular}%
}
\end{table}

%%%%%%%%%%%%%%%%%%%%%%%%%%%%%%%%%%%%%%%%%%%%%%%%%%%%%%%%%%%%%%%%%%%%%%%%%%%%%%%%
\section{Conclusions}
We present here an autonomous driving system which effectively utilizes a learned visual attention model, trained on human gaze data, to improve both training speed and overall performance. Our goal was to demonstrate the usefulness of such models to help solve challenging visuomotor tasks, such as autonomous driving, by transferring the expertise and task knowledge encoded in human eye movements.

To create a system capable of replicating human visual attention, we collected several hours of simulated driving data from human drivers while recording their eye movements. We evaluated four successful published approaches to predicting human gaze on our dataset, and extended the best-performing of these models by providing the high-level intentions of the driver at each frame, boosting performance even further. Our fully-trained model is able to identify key regions in images which are relevant to the driving task, as measured by its high performance in predicting human gaze for an unseen set of test episodes.

Motivated by other works which have improved the performance of machine learning systems by incorporating attention-focusing mechanisms into their learning processes, we tested several methods of introducing this saliency information to an autonomous driving agent. Five driving agents were trained and evaluated, with a substantial reduction in prediction error being attained by a dual-branch model which trains using both raw driving images and attention-masked images. We note the importance of architecture choice when incorporating attention maps into such systems, as our soft agent which in theory had access to the same information as the dual-branch agent did not see any error reduction as a result. Finally, we found that an agent trained solely on the information retained by a hard masking of the original images with the gaze network's attention maps suffered almost no decrease in performance in comparison to a network trained with full driving images, validating the effectiveness of the gaze network in extracting all of the necessary information from these images for the driving task. Furthermore, we believe such an approach, being built on human visual attention, will bring with it further interpretability of the underlying agent -- as it effectively means we now know which parts of the visual input the agent is actually paying attention to, following concepts previously presented in \cite{beyret}.

%To more fully investigate the benefits of the gaze network's predictions on our driving agents, we intend to perform online evaluation of our agents. Metrics such as episode completion rate, average distance between crashes and other driving infractions could give important insight as to the exact effects of the gaze network on our driving agents.

In future works, we would be interested to see the effects of this visual attention model on the performance of Reinforcement Learning approaches to autonomous driving, which have thus far seen limited success due to the sample inefficiency inherent to such a complex state-action space. An attention model could help to address this issue, as the agent should be able to much more rapidly associate highlighted image regions with the reward signals it receives as a result of its actions.


\bibliographystyle{ieeetr}
\bibliography{refs.bib}

\end{document}